\documentclass[runningheads]{llncs}
\usepackage{graphicx}
\usepackage{tikz-dependency}
\usepackage{tikz-qtree}

\usepackage{pifont}
\usepackage{times}
\usepackage{makecell}
\usepackage{booktabs}
\usepackage{url}
\usepackage{caption}
\usepackage{subcaption}
\usepackage{cite}
\usepackage{multirow}
\usepackage{multicol}
\usepackage{amsmath}

\usepackage{amssymb}

\newtheorem{myexample}{Example}[section]

\begin{document}
\title{Syntax-informed Question Answering with Heterogeneous Graph Transformer}
%
%
\author{Fangyi Zhu\inst{1} \and
Lok You Tan\inst{1} \and
See-Kiong Ng\inst{1} \and
Stéphane Bressan\inst{1}}
\authorrunning{F. Author et al.}
%
\institute{National University of Singapore
\email{\{fyzhu,seekiong,steph\}@nus.edu.sg,lok.t@u.nus.edu}
}
\maketitle              
\begin{abstract}
Large neural language models are steadily contributing state-of-the-art performance to question answering and other natural language and information processing tasks. These models are expensive to train. We propose to evaluate whether such pre-trained models can benefit from the addition of explicit linguistics information without requiring retraining from scratch.

We present a linguistics-informed question answering approach that extends and fine-tunes a pre-trained transformer-based neural language model with symbolic knowledge encoded with a heterogeneous graph transformer. We illustrate the approach by the addition of syntactic information in the form of dependency and constituency graphic structures connecting tokens and virtual vertices.

A comparative empirical performance evaluation with BERT as its baseline and with Stanford Question Answering Dataset demonstrates the competitiveness of the proposed approach. We argue, in conclusion and in the light of further results of preliminary experiments, that the approach is extensible to further linguistics information including semantics and pragmatics. 
\end{abstract}
\section{Introduction}

Question answering~\cite{qa} is a field within natural language processing~\cite{jurafskyspeech} that studies the design and implementation of algorithms, tools, and systems for the automatic answering of questions in natural language. Among the many types of question answering~\cite{qalitreview}, this work focuses on extractive question answering. Extractive question answering refers to the task of, given a question and a passage, selecting from the passage a text span corresponding to the answer to the question.

Large language models such as Bidirectional Encoder Representations from Transformers (BERT)~\cite{devlin2019bert} brought competitive performance to many natural language processing tasks, including question answering~\cite{lan_albert_2020}. Although these models are obviously able to learn relevant  linguistic information~\cite{goldberg_assessing_2019}, Kuncoro et al. show that BERT benefits from the addition of syntactic information for various structured prediction tasks~\cite{kuncoro_syntactic_2020}. 

\begin{figure}
    \centering
    \hspace{-10mm}\begin{subfigure}[b]{0.35\columnwidth}
        \centering
     \begin{minipage}[b]{\columnwidth}
    \begin{dependency}
    
    \begin{deptext}[font=\scriptsize, column sep=0.003cm]
     due \& to \& a \& stronger, \& tech \& -oriented \& economy\\
    \end{deptext}
    \depedge{1}{2}{pcomp}
    \depedge{1}{7}{pobj}
    \depedge{7}{3}{det}
    \depedge{7}{4}{amod}
    \depedge{6}{5}{amod}
    \depedge{7}{6}{npadvmod}
    
    
    \end{dependency}
    \caption{A dependency graph.}
    \label{fig:deptree}
    \end{minipage}
    \end{subfigure}
    \begin{subfigure}[b]{0.65\columnwidth}
        \centering
    \begin{minipage}[b]{\columnwidth}
        \begin{tikzpicture}[level distance=20pt, scale=.65]
            \begin{footnotesize}
                \Tree [.NP [.NP \edge[roof]; {transport appliances} ]
                [.PP [.JJ such ]
                [.IN as ]
                [.NP [ [.NP \edge[roof]; {railway locomotives,} ]
                [.NP ships, ]
                [.NP steamboats, ]
                [.CC and ]
                [.NP road vehicles ]
                ]]
                ]]
            \end{footnotesize}
        \end{tikzpicture}
        \caption{A constituency tree.}
        \label{fig:contree}
    \end{minipage}
    \end{subfigure}
    \caption{Examples of syntactic graphic structures.}
    \vspace{-8mm}
\end{figure}
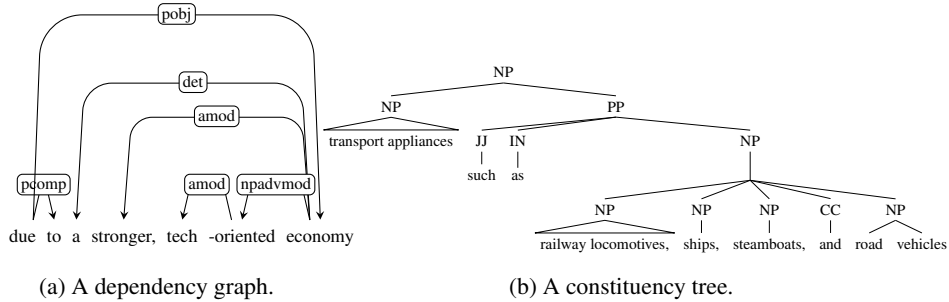

Consider the question ``\textit{What kind of economy did northern California start to grow in the 2000s?}'' from Stanford Question Answering Dataset (SQuAD)~\cite{rajpurkar_know_2018}. The part of the passage in which the answer is located reads ``\textit{[...] due to a stronger tech-oriented economy}''. The answer, according to SQuAD, is ``\textit{tech-oriented}'' (it could also be ``\textit{a  tech-oriented economy}''  or ``\textit{a stronger tech-oriented economy}''). However, BERT is unable to find an answer. A dependency analysis of the sentence, represented by the dependency graph in Figure~\ref{fig:deptree}, reveals that the word ``\textit{tech-oriented}'' is an adverb modifier of ``\textit{economy}''.  This dependency is relevant to the question of the form ``\textit{What kind of [...]}''. Dependencies encode important information specifying grammatical functions between a dependent, here (\textit{tech-oriented}), and its head, here (\textit{economy})~\cite{depparsing}.

Consider the other question ``\textit{Along with road vehicles, locomotives and ships, on what vehicles were steam engines used during the Industrial Revolution?}'', also from SQuAD. The part of the passage in which the answer is located reads ``\textit{[...] Steam engines can be said to have been the moving force behind the Industrial Revolution and saw widespread commercial use driving machinery in factories, mills and mines; powering pumping stations; and propelling transport appliances such as railway locomotives, ships, steamboats and road vehicles. [...]}''. The answer is ``\textit{steamboats}''. BERT finds a redundant answer ``\textit{propelling transport appliances such as railway locomotives, ships, steamboats}''. However, a constituency analysis, represented by the constituency tree in Figure~\ref{fig:contree}, clearly indicates ``\textit{railway locomotives}'', ``\textit{ships}'', ``\textit{streamboats}'', and ``\textit{road vehicles}'' are coordinated noun phrases. 

Generally, the integration of statistical machine learning with symbolic knowledge and reasoning ``opens relevant possibilities towards richer intelligent systems" remark the authors of~\cite{garcez2019neuralsymbolic} arguing for a principled integration of machine learning and reasoning. Nevertheless, most existing neural language models are still plundering the benefits of statistical learning before they attempt to explicitly exploit old-fashion symbolic knowledge of the linguistic structures. 

While the success of transformer-based neural language models is attributed to the self-attention mechanism~\cite{vaswani_attention_2017} that they implement, the question arises whether the adjunction of a focused attention mechanism guided by structures representing symbolic linguistic information~\cite{bergmann2007language}, such as dependency graphs and constituency trees, can further improve the performance of neural language models.

Therefore we devise, present and evaluate a linguistics-informed question answering approach that extends a pre-trained transformer-based neural language model with linguistic graphic structures encoded with a heterogeneous graph transformer~\cite{hgt-www-2020}. The integration is relatively seamless because both models work in the space of embeddings, albeit not necessarily just embeddings of tokens but also of words and other linguistic units. The transformer-based neural language model is fine-tuned and the heterogeneous graph neural network is trained to compute and aggregate the embeddings under the constraints of the graphic structures~\cite{zhou2020graph}.

We instantiate and evaluate the approach for the cases of the addition of syntactic information, in the form of dependency and constituency graphic structures connecting tokens and virtual vertices, for extractive question answering.

We refer to the resulting model as syntax-informed neural network with heterogeneous graph transformer (SyHGT), for dependencies (SyHGT-D) and constituencies (SyHGT-C). For the sake of simplicity, SyHGT is presented, discussed and evaluated here for extractive question answering. 

Overall, there are three main contributions in this work: (1) present a syntax-informed approach via heterogeneous graph transformer for question answering; (2) propose to integrate virtual vertices that can be any linguistic symbolic for incorporating prior linguistic knowledge; (3) empirically evaluate our approach on SQuAD2.0 and it gains 1.22 and 0.98 improvement over the baseline in terms of EM and F1 metrices. 

\section{Related Work}


Early question answering systems used syntactic analysis and rule-based approaches~\cite{baseball}. Later systems utilised heavy feature engineering~\cite{shen-klakow-2006-exploring}.
Advancements in computer hardware then paved the way for neural models which require little feature engineering. 

Language models learn the probability of a sequence of words. Neural language models are often used as encoders to obtain word embeddings. Since BERT, a neural language model, successfully executed on 11 natural language processing tasks, question answering has been dominated by large models built upon it~\cite{lan_albert_2020}.

Jawahar et al. probed BERT's layers, and found that lower layers captured surface features, middle layers captured syntactic features, and upper layers captured semantic features~\cite{jawahar_what_2019}. The upper layers were found to model the long-distance dependencies, making them crucial to performance in downstream tasks. However, it was also found that syntactic information is diluted in these upper layers. Kuncoro et al. extended BERT to take into account syntactic information by modifying its pre-training objective~\cite{kuncoro_syntactic_2020}. Using another syntactic language model as a learning signal, they added what they termed `syntactic bias' to BERT.


\begin{figure*}[!t]
    \centering
  \includegraphics[width=\textwidth]{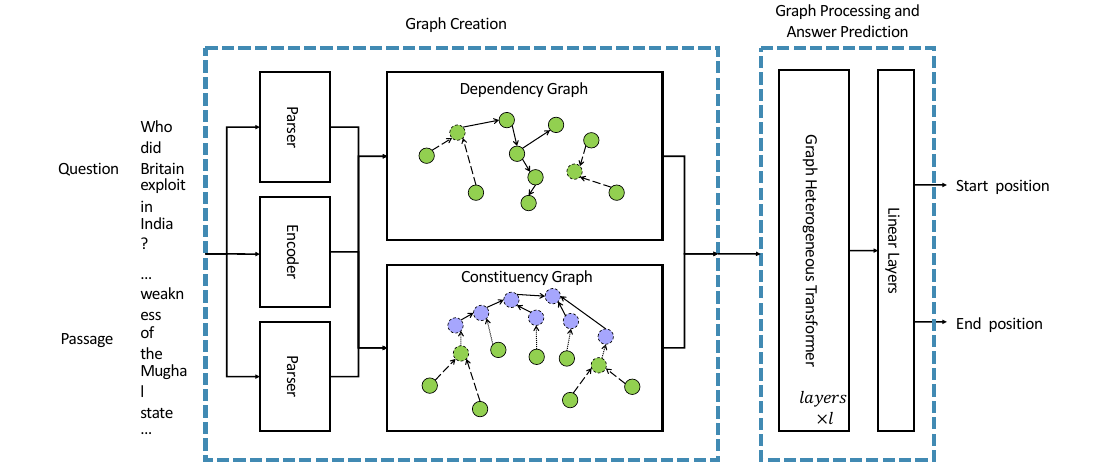}
  \caption{The diagram of our proposed approach. There are two major components, graph creation, as well as graph processing and answer prediction. The green vertices are token vertices, including the dashed green vertices (virtual lexeme vertices).
  The dashed blue vertices in the constituency graph are virtual constituent vertices. The dashed arrows are morphology edges. The straight arrows are dependency edges or constituency edges. The dotted edges in the constituency graph are part-of-speech edges.}
    \vspace{-6mm}
   \label{fig:rgcn}
\end{figure*}

Vashishth et al. used dependency trees and graph convolutional networks to learn syntax-based embeddings that encode functional similarity instead of traditional topical similarity~\cite{vashishth_incorporating_2019}. The syntax-based embeddings were found to encode information complementary to ELMo~\cite{elmo2018} embeddings that only relied on sequential context. Zhang et al. proposed syntax-guided network (SG-Net), a question answering model that used dependency trees as explicit syntactic constraints for a self-attention layer~\cite{zhang_sg-net_2019}. SG-NET was effective especially with longer questions as it could select vital parts. The syntax-guided attention considered syntactic information that is complementary to traditional attention mechanisms. Syntax guidance provided more accurate attentive signals and reduced the impact of noise in long sentences.

For question answering, graph neural networks have found success in multi-hop reasoning~\cite{de_cao_question_2019,tu_multi-hop_2019} on the WikiHop data set~\cite{welbl-etal-2018-construct}. Graph neural networks operate directly on graphs and can capture dependencies between vertices.

This work shows the utility of syntax and graph neural networks in learning better representations. In our approach, we bolster the pre-trained BERT model with additional syntactic information. In the same vein as the approach by Mao et al. ~\cite{nscl}, we bridge old rule-based systems and new neural models by integrating symbolic knowledge into statistical machine learning. This is done by explicitly incorporating the syntactic information, namely constituencies and dependencies, into a question answering model, which is made possible by inserting a heterogeneous graph transformer into the question answering pipeline. This keeps our approach rooted linguistically, instead of solely relying on pre-trained language models that are not explainable. 
Unlike vanilla graph neural networks, a heterogeneous graph transformer can deal with a heterogeneous graph where multiple types of vertices associated with different relations exist. To the best of our knowledge, integrating the syntax information and heterogeneous graph transformers for extractive question answering has not yet been explored.

\section{Methodology}

In a standard neural language model applied to extractive question answering, the question and passage are encoded together, then passed through a linear layer that outputs the probabilities for each token to be the start and end of the answer span.

We propose SyHGT, a linguistics-informed architecture. We need to create, represent, and process linguistic graphic structures connecting the language model embeddings of the tokens of the question and passage. In the cases of syntactic information about dependencies and constituencies, the graphic structures are created by a parser. The result for a question-passage pair is a graph of embedded vectors of the tokens. This non-Euclidean graph structure cannot be used directly by the neural language model. However, the insertion of a heterogeneous graph transformer layer to the question answering pipeline allows us a relatively straightforward implementation combining both the statistical and symbolic information. By inserting the graph neural network between the neural language model and the output layer, we can process the graph before making a prediction. Figure~\ref{fig:rgcn} depicts, SyHGT, the proposed approach, with its two main components: graph creation and graph processing and answer predication. 

\subsection{Graph Creation Module}
Both passage and question are parsed and encoded. The encoder produces embeddings for each token that correspond to the graph vertices. The syntax graphic structures define  the graph vertices and edges. Respectively, we create a constituency graph and a dependency tree. The obtained graphs are the input to the following heterogeneous graph transformer. Note that the tokeniser of the neural language model may not align with the tokeniser of the parser, be it for dependencies or for constituencies, which most likely considers lexemes rather than morphemes. The graph heterogeneous graph neural network model  easily alleviates this issue by the introduction of intermediary vertices aggregating tokens into lexemes, where needed.

\vspace{-4mm}
\subsubsection{Encoder}
SyHGT requires a neural language model as its initial encoder which produces embedding vectors for the text. The Bidirectional Encoder Representations from Transformers (BERT), see Vaswani et al.~\cite{transformers} and Devlin et al.~\cite{devlin2019bert}, is used for implementation and performance evaluation in this paper. 

The question $q$ and passage $p$ are concatenated with the appropriate BERT-specific special tokens to form the sequence: [CLS] $q$ [SEP] $p$ [SEP]. The sequence is fed into BERT to obtain the token embeddings $T = {t_{1}, ..., t_{n}}$, which are the hidden states of the input sequence at the last layer, where $n$ is the number of tokens.

\vspace{-4mm}
\subsubsection{Dependency graph}
\label{sec:dependency-graph}
Dependency is the notion that linguistic units, lexemes, e.g., words, are connected to each other by directed links. In a dependency structure, every lexical vertex is dependent on at one other lexical vertex or is the head of a dependency. The structure is therefore a directed graph, with vertices representing lexical elements and edges representing dependency relations~\cite{nivre-2003-efficient}. Dependency parsing produces a dependency graph of a sentence. The dependency graphs are then processed to obtain the dependency relations. Each sentence in the question and passage is parsed individually. 

Since most BERT implementations leverage the WordPiece tokenizer~\cite{wu2016google}, which may split words into sub-words, i.e. ad hoc morphemes, we add, in such a case, a virtual lexeme vertex on top of the sub-word tokens to represent the original word, so that the graph construction happens at the correct level. 

The edges in the dependency graph are grouped into two categories morphology edges and dependency edges. Morphology edges connect token vertices corresponding to sub-words to virtual lexeme vertices. Dependency edges connect the head vertex and the dependent vertex of a recognized dependency relation. There is one type of edge for each type of dependency relation.

\vspace{-4mm}
\subsubsection{Constituency graph}
\label{sec:constituency-graph}
Constituency analysis iteratively decomposes sentences into constituent or sub-phrases, which are clauses, phrases, and words. These constituents belong to one of several  categories such as noun phrase (NP), verb phrase (VP) as well as parts of speech. Explicitly, given an input sentence, constituency analysis builds a tree, in which leaves or terminal vertices correspond to input words and the internal or non-terminal vertices are constituents.

The vertices in the constituency tree are grouped into three categories, token vertices, lexeme vertices, and constituent vertices. Token vertices correspond to common tokens. Lexeme vertices represent lexemes that need to be recomposed from the token vertices of their sub-words. Constituent vertices represent constituents. Inner nodes and the root of the constituency tree are virtual, lexeme or constituency, vertices.

The edges in the constituency tree are grouped into three categories, morphology edges, part-of-speech edges, and constituent edges.
Morphology edges connect token vertices corresponding to subwords to lexeme vertices. Part-of-speech edges connect the part-of-speech vertices to lexemes vertices. Constituency edges connect low-level constituents to high-level constituents.

\subsection{Graph Processing and Answer Prediction}
The heterogeneous graph transformer takes the constructed graphs as input and passes its outputs to the linear layer. The output from the linear layer consists of two numbers for each vertex; one number denotes the probability of the vertex being the start of the answer span, and the other of the vertex being the end. The final predicted start position and end position of span is determined by the respective maximum scores.

\vspace{-6mm}
\subsubsection{Heterogeneous Graph Transformer}
Graph neural networks, proposed by the authors of~\cite{scarselli2009graph}, are neural models that capture the dependence of graphs via message passing following the edges between the vertices in a graph~\cite{wu2020comprehensive,zhou2020graph}. Specifically, the target for a graph neural network layer is to yield a contextualized representations for each vertex via aggregating the information from its surrounding vertices. By stacking multiple layers, the obtained representations of the vertices can be fed into downstream tasks, such as vertex classification, graph classification, link prediction, etc. 

Recent years have witnessed the emerging success of graph neural networks (GNNs) for modeling structured data. However, most GNNs are designed for homogeneous graphs, in which all vertices and edges belong to the same types, making them infeasible to represent heterogeneous structures~\cite{hgt-www-2020}. Relational Graph Convolutional Network (R-GCN) first proposed relation-specific transformation in the message passing steps to deal with various relations~\cite{michael2018modeling}. Subsequently, several  works focused on dealing with the heterogeneous graph~\cite{heterogeneous-kdd-2019, wang2019heterogeneous}.
Inspired by the architecture design of Transformer~\cite{vaswani_attention_2017}, Hu et al.~\cite{hgt-www-2020} presented the Heterogeneous Graph Transformer that incorporates the self-attention mechanism in a general graph neural network structure that can deal with a heterogeneous graph. 

Given a heterogeneous graph $G=(V,E)$, each vertex $v\in V$ and each edge $e\in E$ are associated with their type $c\in C$ and $r \in R$. The process in one heterogeneous graph transformer layer can be decomposed into three steps: heterogeneous mutual attention calculation, heterogeneous message passing, as well as target-specific aggregation.

\vspace{-4mm}
\paragraph{Heterogeneous mutual attention calculation}
For a source vertex $s$ of type $c_s$ and a target vertex $t$ of type $c_t$ connected by an edge $e=(s,t)$ of type $r_{e}$,
we first calculate a query vector $Q_{t}$ and a key vector $K_{s}$, with the output from previous heterogeneous graph transformer layer, by two vertex type-specific linear projections $W^{Q}_{c_{t}}$ and $W^{K}_{c_{s}}$,
\begin{align}
Q_{t} &=W^{Q}_{c_{t}}h^{(l-1)}_{t},\\
K_{s} &=W^{K}_{c_{s}}h^{(l-1)}_{s},
\label{eq:q-k-computation}
\end{align}
here $h^{(l-1)}_{s}$ and $h^{(l-1)}_{t}$ denote the representations of vertex  $s$ and vertex $t$ by the $(l-1)$-th heterogeneous transformer layer, separately.

Then, we calculate a similarity score by taking the dot product of $Q_{t}$ with $K_{s}$ as shown in Equation~\eqref{eq:attention-computation}. An edge type-specific linear projection $W^{A}_{r_{e}}$ is utilised in case that there are multiple types of edges between a same vertex type pair, while $\mu$ is a predefined vector indicating the general significance of each edge type. The obtained score is normalised by the square root of the dimension of key vector $d_{K_{s}}$. After the scores for all neighbors of $t$ have been computed, a softmax function is applied to yield the normalised attention weights $A_{t}$ for neighbor aggregation,

\begin{equation}
A_{t} =\underset{\forall s \in N_{t}}{softmax}(\frac{\mu K_{s}W^{A}_{r_{e}}Q_{t}^{T}}{\sqrt{d_{K_{s}}}}). \label{eq:attention-computation}
\end{equation}

\vspace{-3mm}
\paragraph{Heterogeneous message passing}
Parallel to the mutual attention calculation, the representation of source vertex $s$ from previous heterogeneous graph transformer layer $h^{(l-1)}_{s}$, is fed into another linear projection $W_{c_{s}}^{M}$ to produce a message vector $M_{s}$,
\begin{align}
    M_{s} &=W_{c_{s}}^{M}h^{(l-1)}_{s}W_{r_{e}}^{M},
\end{align}
here we add another projection $W_{r_{e}}^{M}$to incorporate the edge dependency.

\vspace{-2mm}
\paragraph{Target-specific aggregation}
With the attention weights $A_{t}$ and message vector $M_{s}$ yielded by previous steps,  we aggregate the information from all the neighbors to $t$,
\begin{equation}
h^{(l)}_{t} = \sigma (W^{C}_{c_{t}}\sum\limits_{s\in N_{t}}A_{t}M_{s})+h^{(l-1)}_{t},
\end{equation}
where $W^{C}_{c_{t}}$ is another linear projection mapping the aggregated representation back to $t$'s type-specific feature space, followed by a non-linear activation operation. By conducting the residual connection operation~\cite{he2016residual}, a highly contextualized representation $h^{(l)}_{t}$ for the target vertex $t$ by the current $l$-th heterogeneous graph transformer layer is produced that can be fed into the following module for downstream tasks.

\vspace{-3mm}
\subsubsection{Answer Prediction}
After propagation by the heterogeneous graph transformer layers, the produced representations for the vertices corresponding to common tokens $h$ are passed to the linear projections $W_{s}$ and $W_{e}$ ,
\begin{align}
    y_{s} &= softmax(W_{s}h),\\
    y_{e} &= softmax(W_{e}h).
\end{align}
The two probability distributions $y_{s}$ and $y_{e}$ indicate the probability of each vertex being the start or end of the answer span separately. 

We compute the cross entropy loss as our training objective,
\begin{equation}
    \mathcal{L} = - (y_{s}^{\prime}\log y_{s} + y_{e}^{\prime} \log y_{e}),\\
\end{equation}
where $y_{s}^{\prime}$ and $y_{e}^{\prime}$ are the ground truth start position and end position of the answer.

\section{Experiments and Discussions}
We empirically evaluate the effectiveness of our proposed approach with version 2.0 of the Stanford Question Answering Dataset (SQuAD 2.0)~\cite{rajpurkar-etal-2018-know}. 

\vspace{-2mm}
\subsection{Setup}
The base encoder is a pre-trained BERT language model, in its public Pytorch implementation from the \textit{Transformers}\footnote{\url{github.com/huggingface/transformers}} library. We keep their default settings with a maximum input length of 384. We initialise the weights with the saved models available from \textit{Hugging Face}\footnote{\url{huggingface.co/}}. We then fine-tune the weights during training. We use the standard BERT base model (cased), also known as \textit{bert-base-cased}, model. To build the heterogeneous graph transformer, we use the \textit{pytorch-geometric} library.\footnote{\url{github.com/rusty1s/pytorch_geometric}}

In the dependency graph, the embeddings of the initial tokens are obtained from the pre-trained language model. The embeddings for the virtual lexeme tokens are initialised with  the mean of the embeddings of their corresponding sub-words. 
The dependency graph  edges are obtained from  dependency parsing with the method of~\cite{dozat2016deep}. Their embeddings are initialised randomly according to the type of dependency.

In the constituency tree, similarly to the dependency tree, the embeddings of the initial tokens are obtained from the pre-trained language model and the embeddings for the virtual lexeme tokens are  initialised with the mean of the embeddings of their corresponding sub-words. The constituent vertices are obtained from  constituency parsing with the method of~\cite{kitaev-klein-2018-constituency}. The embeddings of the virtual  vertices and of the edges are  initialised randomly according to their category.

The training uses AdamW optimizer~\cite{adamw} and a learning rate of 2e-5. We stack 2 heterogeneous graph transformer layers with 4 attention heads in each. The model is trained with a mini-batch size of 32 for 7 epochs. The code will be available on Github.

Training and testing use SQuAD 2.0, a data set of questions collected on a set of Wikipedia articles. The answer to every question is a text span or the question might be unanswerable. It contains around 130k training and 12k development examples.

\subsection{Evaluation}

\subsubsection{Metrics}
We use the following two metrics for the performance evaluation. F1 measures the normalised average  overlap between the prediction and ground-truth answer. Exact match (EM) evaluates whether the prediction exactly matches the ground-truth.

\vspace{-5mm}
\subsubsection{Overall experimental results}
The overall experimental results are shown in Table~\ref{tab:comparisons-with-baseline}.
We compare the performance of the pre-trained BERT alone, of SyHGT with a dependency graph and BERT, and of SyHGT with a constituency tree and BERT. The results are presented in Table~\ref{tab:comparisons-with-baseline}, in which the three models are refered to as BERT, SyGHT-D (BERT), and SyGHT-C (BERT), respectively. We observe a slight improvement of 0.77 EM and 0.46 F1 of SyGHT-C  over the BERT baseline and a more significant improvement of 1.22 EM and 0.97 F1 of SyGHT-D over the BERT baseline.

\begin{table}[!t]
\centering
\caption{Overall comparative empirical performance with SQuAD2.0.}
\begin{tabular}{lcc}
\toprule
Method & F1 & EM\\ \midrule
BERT & 75.41 & 71.78\\
SyHGT-C (BERT) & 75.87 & 72.55 \\ 
SyHGT-D (BERT) & \textbf{76.38} & \textbf{73.00} \\
\bottomrule
\end{tabular}
\label{tab:comparisons-with-baseline}
\vspace{-8mm}
\end{table}

\subsubsection{Microanalysis}
\paragraph{Dependency Graph}
We examine the dependencies\footnote{The descriptions of the dependencies can be found in \url{downloads.cs.stanford.edu/nlp/software/dependencies_manual.pdf}.}  involved in question-answer pairs for which SyGHT-D and BERT alone find different answers. Each example shows the question (Q), the paragraph (P) as well as the BERT and SyHGT-D answers.

Examples~\ref{ex:D1} and ~\ref{ex:D2}, along with the corresponding dependency graphs, showcase the inferred utility of dependencies. 

\vspace{-2mm}
\begin{small}

\hspace{-5mm}\begin{minipage}[t]{0.48\columnwidth}
\begin{myexample}
\label{ex:D1}\hfill

\textit{Q: Colonial rule would be considered what type of imperialism?}

\textit{P: ... Formal imperialism is defined as physical control or full-fledged colonial rule ...}

BERT: \textit{Formal imperialism is defined as physical control or full-fledged} 

SyHGT-D: \textit{Formal}
\vspace{-5mm}
\begin{center}
\noindent\begin{dependency}
\begin{deptext}[font=\small,column sep=-0.05cm]
\hspace{-5mm} Formal \& imperialism \& is \& defined \& as \& physical \& control \\\end{deptext}
\depedge{2}{1}{amod}
\depedge{4}{2}{nsubpass}
\depedge{4}{3}{auxpass}
\depedge{4}{5}{prep}
\depedge{5}{7}{pobj}
\depedge{7}{6}{amod}
\end{dependency}
\end{center}
\end{myexample}
\end{minipage}
\hspace{3mm}\begin{minipage}[t]{0.48\columnwidth}
\begin{myexample}
\label{ex:D2}\hfill

\textit{Q: What is the other country the Rhine separates Switzerland to?}

\textit{P: The Alpine Rhine begins ... and later forms the border between Switzerland to the West and Liechtenstein ...}
\\

BERT: \textit{the West and Liechtenstein}

SyHGT-D: \textit{Liechtenstein}
\\
\vspace{-9mm}
\begin{center}
\begin{dependency}
\begin{deptext}[font=\small, column sep=0.05cm]
Switzerland \& to \& the \& West \& and \& Liechtenstein \\\end{deptext}
\depedge{1}{2}{prep}
\depedge[edge unit distance=2.48ex]{2}{4}{pobj}
\depedge{4}{3}{det}
\depedge[edge unit distance=1.8ex]{1}{5}{cc}
\depedge[edge unit distance=1.7ex]{1}{6}{conj}
\end{dependency}
\end{center}
\end{myexample}
\end{minipage}
\end{small}

In Example~\ref{ex:D1}, BERT predicts a long and incorrect span whereas SyHGT-D, informed by the dependency graph,  recognises that  `formal', as an adjectival modifier (\textit{amod}) of `imperialism', is the correct answer.
In Example~\ref{ex:D2}, the dependency tree shows that the phrase `to the West' is connected to `Switzerland' through a preposition (\textit{prep}) as an object of preposition (\textit{pobj}) , while `Liechtenstein' is a conjunct (\textit{conj}). SyHGT-D uses the dependencies to correctly identify that `to the West' is not a separate element from `Switzerland', and that `Liechtenstein' is the answer.

Overall, we report  that several specific dependencies, in particular \textit{prep, pobj, dobj, nsubj, conj, cc} seem to allow SyHGT-D to predict corresponding linguistically sound answers, albeit sometimes at the expense of the more general answer. SyHGT-D seems to be parsimonious. 

\paragraph{Constituency Graph}
We examine the constituencies\footnote{The descriptions of the dependencies can be found in \url{http://surdeanu.cs.arizona.edu//mihai/teaching/ista555-fall13/readings/PennTreebankConstituents.html}.}  involved in question-answer pairs for which SyGHT-C and BERT alone find different answers. Each example shows the question (Q), the paragraph (P) as well as the BERT and SyHGT-C answers.

Examples~\ref{ex:C1} to~\ref{ex:C4}, along with the corresponding constituency trees, showcase the inferred utility of dependencies.

\begin{small}

\hspace{-5mm}\begin{minipage}[t]{0.48\columnwidth}
\begin{myexample}\hfill

\label{ex:C1}
\textit{Q: What is one example of what a clinical pharmacist's duties entail?}

\textit{P: ... The clinical pharmacist's role involves creating a comprehensive drug therapy plan for patient-specific problems, identifying goals of therapy, and reviewing ...}
\\

BERT: \textit{creating a comprehensive drug therapy plan for patient-specific problems, identifying goals of therapy, and reviewing all prescribed medications} 

SyHGT-C: \textit{creating a comprehensive drug therapy plan for patient-specific problems}
\\

\hspace{-5mm}\begin{tikzpicture}[level distance=20pt, scale=.7]
    \begin{footnotesize}
        \Tree [.S [.NP \edge[roof]; {The ... role} ]
        [.VP [.VBZ involves ]
        [.VP \edge[roof]; {creating ... problems,} ]
        [.VP \edge[roof]; {identifying ... therapy,} ]
        [.CC and ]
        [.VP \edge[roof]; {reviewing ...} ]
        ] ]
    \end{footnotesize}
\end{tikzpicture}
\end{myexample}

\end{minipage}
\hspace{3mm}\begin{minipage}[t]{0.48\columnwidth}

\begin{myexample}
\label{ex:C2}\hfill

\textit{Q: Where did Kublai extend the Grand Canal to?}

\textit{P: ... Kublai expanded the Grand Canal from southern China to Daidu in the north. ...}
\\

BERT: \textit{southern China to Daidu in the north} 

SyHGT-C: \textit{Daidu in the north}
\\

\hspace{-3mm}\begin{tikzpicture}[level distance=20pt, scale=.7]
    \begin{footnotesize}
        \Tree [.S [.NP [.NNP Kublai ] ]
        [.VP [.VBD expanded ]
        [.NP \edge[roof]; {the Grand Canal} ] 
        [.PP [.IN from ]
        [.NP \edge[roof]; {southern China} ]]
        [.PP [.TO to ]
        [.NP \edge[roof]; {Daidu in the north.} ]]
        ] ]
    \end{footnotesize}
\end{tikzpicture}
\end{myexample}
\end{minipage}
\end{small}

\begin{small}

\hspace{-5mm}\begin{minipage}[t]{0.48\columnwidth}

\begin{myexample}
\label{ex:C3}\hfill

\textit{Q: Who published the State of the Planet 2008-2009 report?}

\noindent\textit{P: Michael Oppenheimer, a long-time participant in the IPCC and coordinating lead author of the Fifth Assessment Report conceded in Science Magazine's State of the Planet 2008-2009 ...}
\\

\noindent BERT: \textit{Michael Oppenheimer} 

\noindent SyHGT-C: \textit{Science Magazine}
\\

\hspace{-8mm}\begin{tikzpicture}[level distance=20pt, scale=.7]
    \begin{footnotesize}
        \Tree [.VP [.VBD conceded ]
        [.PP [.IN in ]
        [.NP [.NP [.NNP Science ] 
        [.NN Maganize ] 
        [.POS \text{'}s ]]
        [.NAC \edge[roof]; {State pf the Planet} ]
        [.CD 2008 ]
        [.\text{:} \text{-} ] 
        [.CD 2009 ]
        ]]]
    \end{footnotesize}
\end{tikzpicture}
\end{myexample}
\end{minipage}
\hspace{3mm}\begin{minipage}[t]{0.48\columnwidth}
\begin{myexample}
\label{ex:C4}\hfill

\textit{Q: What type of architecture is represented in the majestic churches?}

\textit{P: Gothic architecture is represented in the majestic churches but also at the burgher houses and fortifications. ...}

BERT: \textit{Gothic architecture} 

SyHGT-C: \textit{Gothic}

\hspace{-13mm}\begin{tikzpicture}[level distance=20pt, scale=.7]
    \begin{footnotesize}
        \Tree [.S
        [.NP [.JJ Gothic ]
        [.NN architecture ]]
        [.VP [.VBZ is ]
        [.VP [.VBN represented ]
        [.PP [.PP \edge[roof]; {in the majestic} ]
        [.CC but ]
        [.RB also ]
        [.PP but \edge[roof]; {at the ...} ]]]]]
    \end{footnotesize}
\end{tikzpicture}
\end{myexample}
\end{minipage}
\end{small}

In Example~\ref{ex:C1}, the question asks for one clinical pharmacist's duty where the answer is one of `creating a comprehensive drug therapy plan for patient-specific problems', `identifying goals of therapy', `reviewing ...'. BERT is not able to distinguish different duties based on the syntax structure and thus gives all the duties in the passage. However, the constituency tree helps distinguish the three duties. Hence, with the constituencies, SyHGT-C can provide a correct answer.
In Example~\ref{ex:C2}, BERT predicts `southern China to Daidu in the north. BERT  confuses the coordinated prepositional phrases (PP) `from southern China' and `to Daidu in the north'. With the constituency tree, SyHGT-C can understand that `Kublai expanded' from the start location `southern China' to the end location `Daidu in the north' and predict the correct answer.
The passage in Example~\ref{ex:C3} is difficult to understand as it contains long and complex clauses. BERT fails to understand this sentence.  The constituency tree clarifies that `State of the Planet' belongs to `Science Magazine' leading to the correct answer. Example~\ref{ex:C4} further illustrates that SyHGT-C can give  more accurate answers that exactly match the ground-truth benefited from integrating the constituency information.
\vspace{-1mm}

\subsection{LingHGT, SemHGT, PragHGT: Towards linguistics-informed language models}

Linguistics structures are numerous and, more often than not, amenable to a graph representation. Such structures exist not only for many aspects of syntax but also for semantics and pragmatics. The architecture we discussed applies to other linguistic graphic structures.
Preliminary experiments with BERT and SQuAD seem to confirm the versatility and effectiveness of the model and its realisation for lingustics in general, LingHGT, and for semantics, SemHGT, and pragmatics, PragHGT, in particular.

We are conducting preliminary experiments of the utilisation of an entity-relationship graph SQuAD for semantics- and pragmatics-informed question answering.  We used the spaCy library~\cite{spacy}\footnote{\url{https://spacy.io/}} for named entity recognition and the OpenNRE~\cite{han-etal-2019-opennre}\footnote{\url{https://github.com/thunlp/OpenNRE}} for relationship extraction. The model is pre-trained on the the Wiki80 data set that is derived from FewRel and covers 80 relations~\cite{han-etal-2018-fewrel}\footnote{\url{https://github.com/thunlp/FewRel}}. We extract entities and relationships from the questions and passages for semantic information and we augment questions with contextual information about the questioner for pragmatic information.
In the following, we look at examples where BERT incorporating semantics information can answer correctly while the original BERT cannot. 
In Example~\ref{ex:S2}, the relevant relation identified is `child' between the entity `Lupe Mayorga' (\textit{PERSON}), and the entity `Aken' (\textit{PERSON}). 
In Example~\ref{ex:S3} the relevant relation identified is `member of political party' between the entity `Annabel Goldie' (\textit{PERSON}), and the entity `Conservatives' (\textit{NORP}). 







\begin{small}
\hspace{-5mm}\begin{minipage}[t]{0.48\columnwidth}
    
\begin{myexample}\hfill
\label{ex:S2}

\textit{Q: Who was Bill Aiken's adopted mother?}

\textit{P: Aken, adopted by Mexican movie actress Lupe Mayorga}

BERT: \textit{\textlangle no answer\textrangle}

SemHGT:  \textit{Lupe Mayorga} 
\end{myexample}
\end{minipage}
\hspace{3mm}\begin{minipage}[t]{0.48\columnwidth}
\begin{myexample}\hfill
\label{ex:S3}

\textit{Q: Who announced she would step down as leader of the Conservatives?}

\textit{P: ...leader Annabel Goldie claiming that their support had held firm. Nevertheless, she too announced she would step down as leader of the party...}

BERT: \textit{\textlangle no answer\textrangle}

SemHGT:  \textit{Annabel Goldie} 
\end{myexample}
\end{minipage}
\end{small}

We are exploring the opportunity and the applications of a pragmatics-informed language model. The following example is simulated in order to illustrate the targetted behaviour of a pragmatics-informed version of the proposed model. 

\vspace{-2mm}
\begin{small}
\begin{myexample}\hfill
\label{ex:P1}


\textit{P: The IPCC receives funding through the IPCC Trust Fund, established in 1989 by the United Nations Environment Programme (UNEP) and the World Meteorological Organization (WMO), Costs of the Secretary and of housing the secretariat are provided by the WMO, while UNEP meets the cost of the Depute Secretary.}

\textit{Q (asked by the Secretary) :Who funds my secretariat?} 

PraHGT: \textit{the World Meteorological Organization}

\textit{Q (asked by the Deputy Secretary) :Who funds my secretariat?}

PraHGT: \textit{the United Nations Environment Programme}
\end{myexample}
\vspace{-3mm}
\end{small}

In Example~\ref{ex:P1}, PraHGT should leverage its knowledge of the classes of the entities `The United Nations Environment Programme' and `the World Meteorological Organization', namely `ORG - organization', `Secretary' and `Deputy Secretary', `PER - person', and the relationships connecting them directly or via other entities to the questioner to produce correct answers. The spaCy named-entity recogniser does not have a class `Job Title' for `Secretary' and `Deputy Secretary'. The custom class needs to be added. The reader notices that a morphology informed tokenisation is also needed in order to guarantee the proper association of `secretariat' with 'Secretary', and to understand the typographical error (original to SQuaD2.0) in 'Depute' (instead of `Deputy'.)

In general we believe that LingHGT is a blueprint for the implementation of linguistics-informed models on top of the existing powerful pre-trained neural language models, wherever the linguistics information can be represented as a graph.

\section{Conclusion}
This paper presented a syntax-informed question answering model. The approach combines the statistical knowledge of neural language model with the symbolic information contained in linguistic graphic structures such as dependencies graphs and constituency trees. The seamless integration is realised by the means of a heterogeneous graph transformer added to a pre-trained transformer-based neural language model. The models therefore combines the self-attention mechanism of the transformer-based neural language model with a focused attention guided by graphic structures representing linguistics information in a heterogeneous graph transformer model. 

An empirical performance evaluation of the proposed approach in comparison to the neural language model alone for question answering with SQuAD2.0 shows improvement. An initial microanalysis of the results suggest that the proposed model makes more focused predictions thanks to its awareness of syntax. Several examples, for which the proposed approach does not find the correct answer, even suggest that a better syntax parser could be key to addressing the shortcomings. 

Preliminary results for LingHGT, SemHGT, and PragHGT confirm the versatility and effectiveness of linguistics-informed language models and give a blueprint for the implementation of incorporating the linguistics information as a graph into the powerful transformed-based language models.

\section*{Acknowledgements}

This research is supported by the National Research Foundation, Singapore under its Industry Alignment Fund – Pre-positioning (IAF-PP) Funding Initiative. Any opinions, findings and conclusions or recommendations expressed in this material are those of the authors and do not reflect the views of National Research Foundation, Singapore.

%
%
%
\bibliographystyle{splncs04}
\bibliography{anthology,custom,bibliography}

\end{document}